\def\year{2022}\relax
\documentclass[letterpaper]{article} % DO NOT CHANGE THIS
\usepackage{aaai22}  % DO NOT CHANGE THIS
\usepackage{times}  % DO NOT CHANGE THIS
\usepackage{helvet}  % DO NOT CHANGE THIS
\usepackage{courier}  % DO NOT CHANGE THIS
\usepackage[hyphens]{url}  % DO NOT CHANGE THIS
\usepackage{graphicx} % DO NOT CHANGE THIS
\usepackage{natbib}  % DO NOT CHANGE THIS AND DO NOT ADD ANY OPTIONS TO IT
\usepackage{caption} % DO NOT CHANGE THIS AND DO NOT ADD ANY OPTIONS TO IT
\usepackage{amsfonts}
\usepackage{amsmath}
\usepackage{booktabs}
\DeclareCaptionStyle{ruled}{labelfont=normalfont,labelsep=colon,strut=off} % DO NOT CHANGE THIS
\usepackage{algorithm}
\usepackage{algorithmic}

\usepackage{newfloat}
\usepackage{listings}
\floatstyle{ruled}
\newfloat{listing}{tb}{lst}{}
\floatname{listing}{Listing}
\title{Unsupervised Coherent Video Cartoonization with Perceptual Motion Consistency}
\author{
    Zhenhuan Liu\textsuperscript{\rm 1,2},
    Liang Li\textsuperscript{\rm 1}\thanks{Corresponding author.},
    Huajie	Jiang\textsuperscript{\rm 3} \\
    Xin	Jin\textsuperscript{\rm 3},
    Dandan Tu\textsuperscript{\rm 3},
    Shuhui Wang\textsuperscript{\rm 1},
    Zheng-Jun Zha\textsuperscript{\rm 4}
}
\affiliations {
    \textsuperscript{\rm 1} Institute of Computing Technology, \textsuperscript{\rm 2} University of Chinese Academy of Sciences\\
    \textsuperscript{\rm 3} Huawei Technologies,  \textsuperscript{\rm 4} University of Science and Technology of China \\
    zhenhuan.liu@vipl.ict.ac.cn, liang.li@ict.ac.cn, \{jianghuajie1, jinxin11, tudandan\}@huawei.com, \\ wangshuhui@ict.ac.cn, zhazj@ustc.edu.cn
}
\begin{document}
\maketitle

\begin{abstract}
    % 背景
    In recent years, creative content generations like style transfer and neural photo editing have attracted more and more attention. Among these, cartoonization of real-world scenes has promising applications in entertainment and industry. Different from image translations focusing on improving the style effect of generated images, video cartoonization has additional requirements on the temporal consistency. In this paper, we propose a spatially-adaptive semantic alignment framework with perceptual motion consistency for coherent video cartoonization in an unsupervised manner. The semantic alignment module is designed to restore deformation of semantic structure caused by spatial information lost in the encoder-decoder architecture. Furthermore, we devise the spatio-temporal correlative map as a style-independent, global-aware regularization on the perceptual motion consistency. Deriving from similarity measurement of high-level features in photo and cartoon frames, it captures global semantic information beyond raw pixel-value in optical flow. Besides, the similarity measurement disentangles temporal relationships from domain-specific style properties, which helps regularize the temporal consistency without hurting style effects of cartoon images. Qualitative and quantitative experiments demonstrate our method is able to generate highly stylistic and temporal consistent cartoon videos.
\end{abstract}

\section{Introduction}
\begin{figure}[t]
    \includegraphics[width=\linewidth]{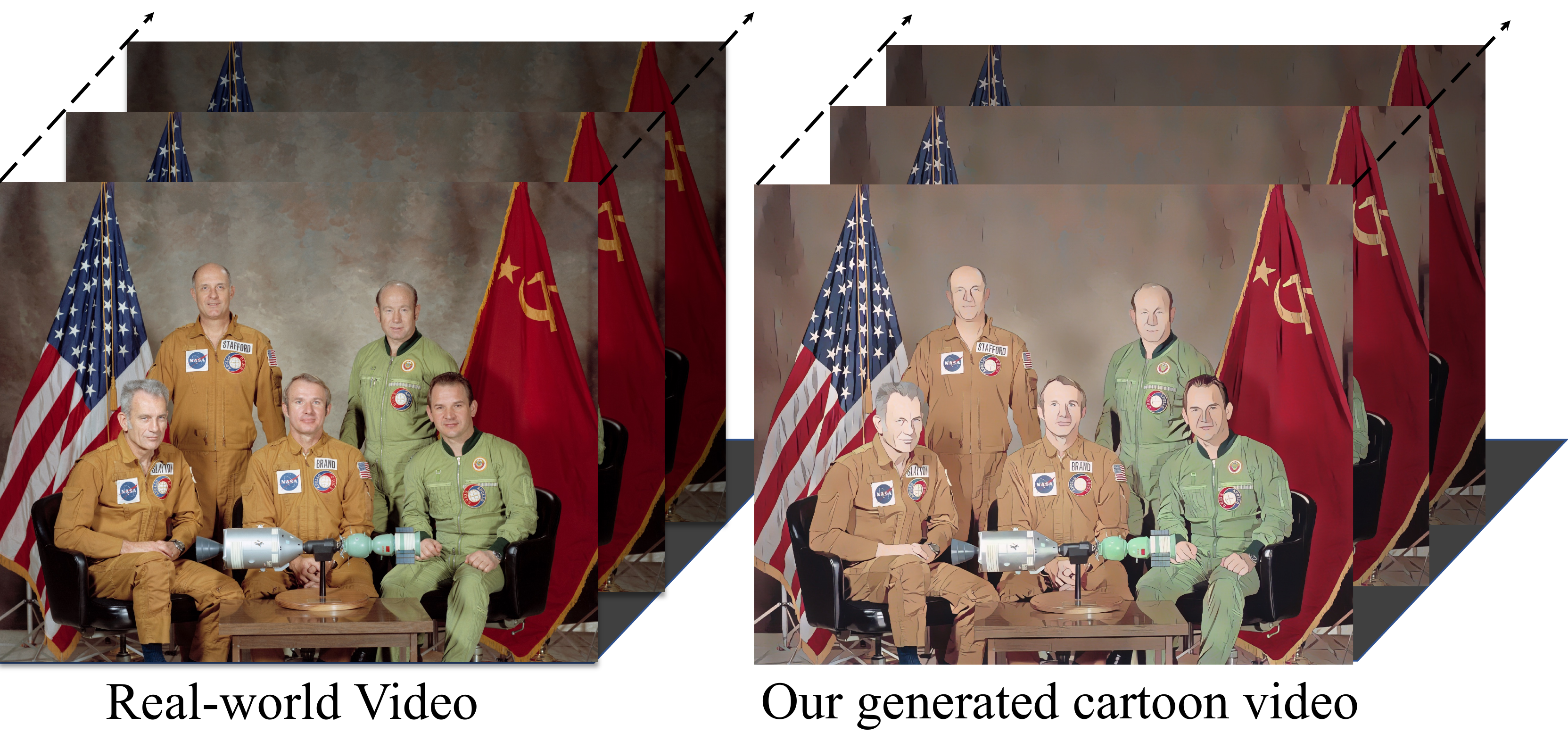}
    \caption{
        Illustration of video cartoonization. It requires smooth surface and clear border with temporal consistency for a typical cartoon artwork.
    }
    \label{fig:intro}
\end{figure}

Cartoon movies are very popular and attractive across the world,
but creating even a short cartoon movie involves a complex and time-consuming production process with multiple stages.
Recently, deep learning based methods have developed a lot of techniques for creative content generation,
like style transfer, semantic image synthesis and neural photo editing.
Video cartoonization aims at creating coherent cartoon-styled videos based on real-world videos, as shown in Figure~\ref{fig:intro}.
As a practical technique, it has promising applications in entertainment and industry.

% image cartoonization
Given two collections containing unpaired photo and cartoon images,
cartoonization usually leverages generative adversarial networks(GANs) to map the input photo into the distribution of cartoon images.
Different from traditional style transfer which often adds textures like brush strokes,
cartoonization aims to create highly conceptual and abstract images.
Besides, due to the dynamic training characteristic of unsupervised GANs,
it's harder to achieve stable convergence of style effects in generated images than style transfer which has explicit style constraints. %\cite{Gatys2016a}
Compared to image cartoonization, video cartoonization has higher requirements on the temporal consistency between input and output frames.
There are two challenges to the coherence of output video.
The first one is keeping the structure consistency between each input and output frame.
It's difficult to preserve structure consistency of output image while depicting the highly abstracted cartoon image structure in unsupervised learning.
The second challenge is to guarantee the temporal coherence of generated frames.
Previous works have shown that the output of deep neural networks is very sensitive to small changes in pixel values ~\cite{Azulay2019}.
The generator of video cartoonization needs to be robust under different transformations of input frames.

\begin{figure*}
    \includegraphics[width=\linewidth]{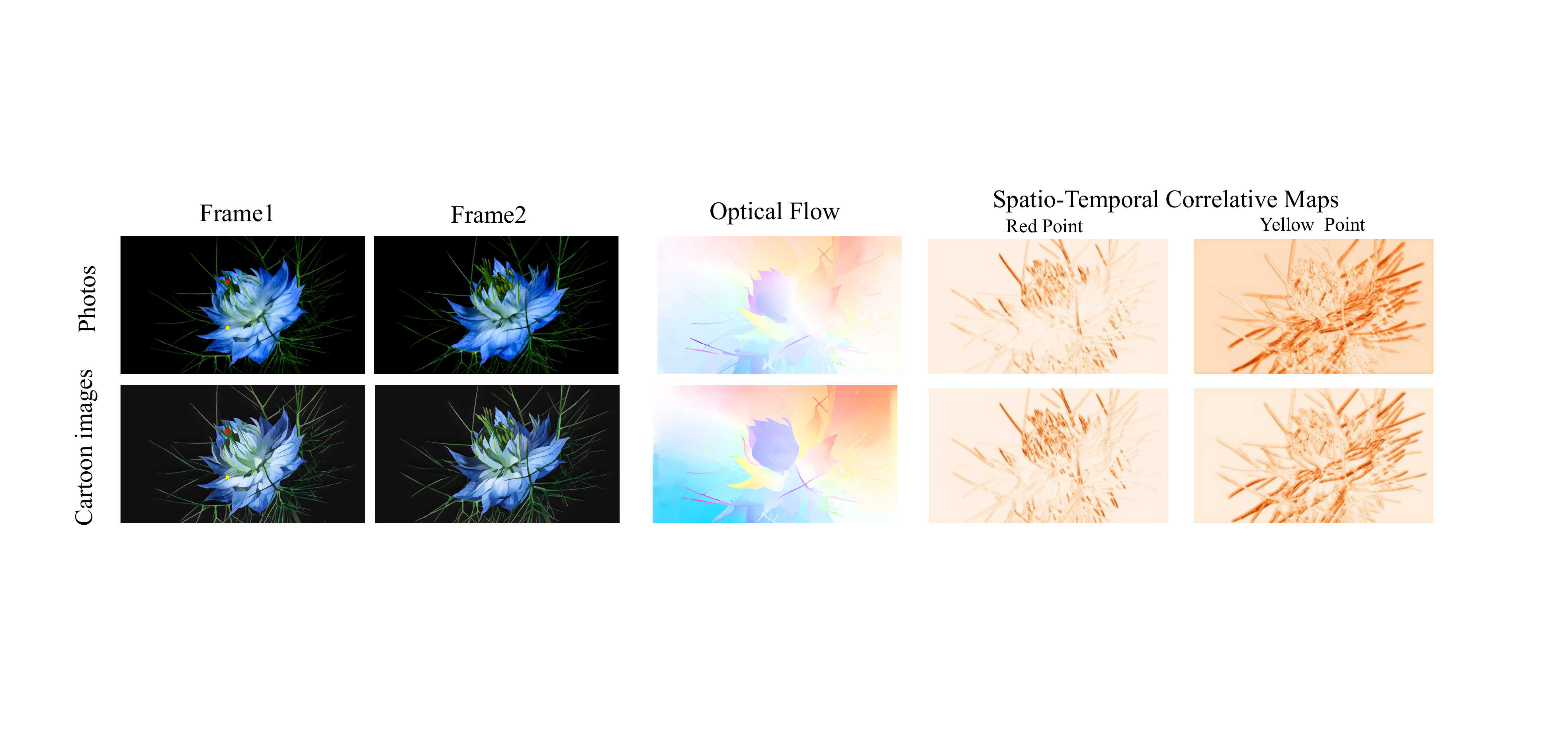}
    \caption{Optical flow and our spatio-temporal correlative maps for two real world photos and its cartoon images.
        We select the red and green points as source points and build their spatio-temporal correlative maps.
        The spatio-temporal correlative maps are more robust under different domains.}
    \label{fig:visual}
\end{figure*}
There are mainly three kinds of methods to solve the problem of video temporal inconsistency.
The first method explicitly employs optical flow estimated between input frames to warp the last output frame for generation~\cite{Chen2017b}.
However, this method highly relies on the accuracy of flow estimation, which is difficult for complex real-world environments.
The second one is to build a task-independent model that can repair temporal inconsistency of videos generated by different image-translation models, such as Deep Video Prior\cite{Lei2020}.
But this method is time-consuming and hard for real-time applications.
Another method is to add a temporal regularization upon image-to-image translation models at training stage,
which encourages the coherence of output frames~\cite{Wang2020b}.
% This method usually smooths regions with notable textures and degrades the stylistic effect.
These traditional methods for temporal consistency often suffer from the limitations of pixel-wise motion representation, e.g., optical flow.
Since images in different domains have different properties in appearances and textures,
optical flow estimated in the photo domain may not align well with the cartoon domain.
As shown in the third column of Figure~\ref{fig:visual}, the optical flows estimated for real images and cartoon images
show obvious discrepancies, especially for the central region with large distortions.
Further, pixel-wise motion representation is unable to handle disocclusion problems where newly appeared pixels have no correspondence with last frame.

% 引出各个模块, 和他们的contribution
To solve above problems, we propose a spatially-adaptive semantic alignment framework
with perceptual motion consistency for coherent video cartoonization in an unsupervised manner.
The Spatially-adaptive Semantic Alignment(SSA) module is designed to restore the local shift and deformation of semantic structure
caused by spatial information lost in the traditional encoder-decoder network architecture.
To overcome the limitations of dense motion representations for complex scenes with large displacements and disocclusion under different domains,
we propose a style-independent global-aware regularization on Perceptual Motion Consistency(PMC) via spatio-temporal correlative maps
between input and output video.

% 细节介绍
Specifically for the SSA module,
we first calculate the semantic distance of feature vectors between encoder and decoder at each scale.
After that, we relocate the decoder's feature in a local patch to the position of its nearest encoder's features,
resulting in a refined structure corresponding to the encoder's structure.
Such that the structure consistency between input and output image can be better preserved.
% 分割线
For the PMC module, we introduce spatio-temporal correlative maps to impose regularization on perceptual motion consistency.
Different from the pixel-wise hard correspondence of optical flow,
it measures feature similarity instead of absolute pixel value for each patch of current frame with next frame.
This map builds a probabilistic distribution of each object's flow direction in a global context,
which can be generalized in complex situations with disocclusion.
Furthermore, the similarity measurement disentangles the temporal relationship from domain-specific attributes such as color, lighting and texture,
so that we can formulate a style-independent motion representation for different domains.

Our contributions can be summarized as follows:
(1) We propose an effective framework with a style-independent global-aware regularization on perceptual motion consistency to generate temporal consistent cartoon videos.
(2) A novel generator with spatially-adaptive semantic alignment is designed to generate semantic-aware structure consistent cartoon images.
(3) We conduct detailed qualitative and quantitative experiments and demonstrate our method achieves both stylistic cartoon effect and temporal consistency. Our code will be released 
on github.
\section{Related Works}

\textbf{Unsupervised Image-to-Image Translation}
aims to learn the mapping from a source image domain to a target image domain with unpaired data.
One of the key challenges is to build a meaningful structure correspondence between input and output images.
Cycle-consistency and its variants are often used to guarantee the output image can reconstruct the input image~\cite{Zhu2017b,Huang2018b,Liu2017,lin2020multimodal,lin2020learning}.
~\cite{park2020contrastive} incorporated patchwise contrastive learning to achieve structure preservation.
Recently ~\cite{Zheng2021} proposed spatially-correlative maps as domain-invariant representations to restrict structure consistency.
In this paper, we learn the unsupervised translation from photo domain to cartoon domain.
Different from the above methods only applying to images, we extend to generate videos with temporal consistency.

\textbf{Image Cartoonization}
aims to generate cartoon images with clear edges, smooth color shading and relatively simple textures from real-world photos.
Originally, the pioneering work CartoonGAN~\cite{Chen2018a} was proposed for image cartoonization,
which introduced a novel adversarial loss to encourage clear edges.
Recently,~\cite{Wang2020a} developed three white-box image representations that reflect different aspects of cartoon images, including texture, surface and structure.
Although these models can generate cartoon styled videos by applying per-frame translation,
their output videos show temporal inconsistencies and flickering artifacts.

\textbf{Video Temporal Consistency}
is a research topic about solving the flickering problem when applying different kinds of image-based models to videos.
Task-independent methods design a single model for different tasks, which generate a coherent video from separately processed frames.
~\cite{Lai2018a} utilized FlowNet2 to estimate optical flow as temporal loss to train the transformation network.
~\cite{Lei2020} leveraged deep video prior to iteratively optimize each sequence.
% Task-specific
Task-specific approaches develop different strategies according to each domain,
such as designing specific network architectures ~\cite{Tassano2020,Deng2020} or
embedding optical flow estimation to capture information of motion ~\cite{Chen2017b}.
Recently, ~\cite{Wang2020b} proposed compound regularization for temporal consistent video style transfer.
These methods often heavily rely on pixel-wise correspondence which is sensitive to the different appearances of objects in photo and cartoon domain.

\section{Methodology}
\subsection{Overview}
\begin{figure*}[t]
    \centering
    \includegraphics[width=\linewidth]{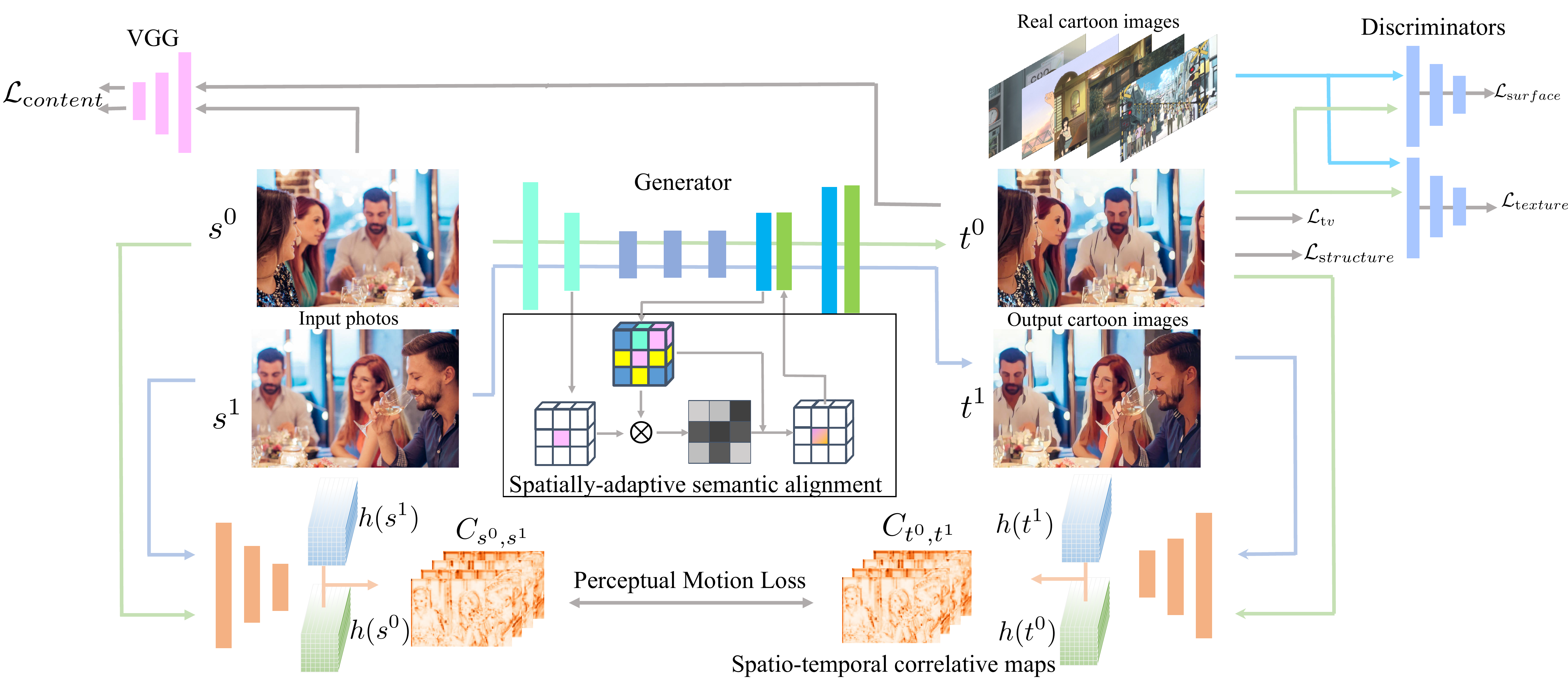}
    \caption{The architecture of our proposed method. Given two consecutive frames of real-world photos,
        the spatially-adaptive semantic alignment module is introduced into our generator to produce semantic-aware structure consistent cartoon images.
        After that we derive the spatio-temporal correlative maps from extracted features of input and cartoon images to regularize the perceptual motion consistency.
        Besides, content loss is used to restrict the structure correspondence between input and output image,
        and adversarial loss is used to optimize the cartoon effect of output images.
    }
    \label{fig:structure}
\end{figure*}

Given a real world video composed of frames $\{s^0, s^1, ...s^n\}$, video cartoonization aims to
generate corresponding cartoon frames $\{t^0, t^1, ...t^n\}$ whose temporal consistency is preserved.
As shown in Figure \ref{fig:structure}, at training stage, our model transforms input consecutive real world photos ${s^0}$ and ${s^1}$ into
corresponding cartoon images $t^0$ and $t^1$.
The generator builds upon an encoder-decoder architecture consisting of downsampling, residual blocks and upsampling layers.
We introduce a spatially-adaptive semantic alignment module into the generator to render semantic-aware structure consistent cartoon images.
To restrict perceptual motion consistency, we first extract multi-level features from both input and cartoon images with pretrained deep network.
Then, we introduce the spatio-temporal correlative maps of features to regularize perceptual motion consistency between input and output video.
We employ adversarial mechanism to optimize the cartoon effect of generated images, where the generator and discriminator are updated alternatively.

\subsection{Spatially-Adaptive Semantic Alignment}
To address the problem of structure deformation in the decoding stage,
we introduce the spatially-adaptive semantic alignment(SSA) module into the generator.
As shown in Figure ~\ref{fig:structure}, to refine the structure of upsampled feature maps $g^{l}(x) \in \mathbb{R}^{C \times H \times W}$ in the $l$-th level of decoder, we adopt the feature maps $f^{l}(x)$ of encoder in corresponding layer to relocate the placement of decoder features.

Specifically, for a source feature vector $f_{i}^{l}(x)$ in location $i$,
we define a local patch of size $N=R^2$ around $i$ in $g^{l}(x)$ as its semantic candidates $\{g_{j}^l(x) | j \in \left\{1,2...,N\right\} \}$.
We first compute the magnitude of similarity between them:
\begin{equation}
    z_{ij}^l(x) = (f_{i}^l(x))^T(g_{j}^l(x))
\end{equation}

Then, we apply spatial-wise softmax to $[z^l_{i1}, z^l_{i2}, ... z^l_{iN}]$, which constructs a normalized correlative map as a kernel.
Each element of the kernel is calculated as follows:
\begin{equation}
    \alpha_{ij}^l(x) = \frac{exp(z_{ij}^l(x))}{\sum_{k=1}^N exp(z_{ik}^l(x))}
\end{equation}

Finally, we obtain the refined feature in location $i$ by aggregating semantic candidates with above kernel weight:
\begin{equation}
    r_{i}^l(x) = \sum_{i=1}^N \alpha_{ij}^l(x) g_{j}^l(x)
\end{equation}

% 性质
Similar to pooling operations, this module replaces the output of convolution with a summary statistic of the nearby outputs.
Furthermore, there are two properties of our proposed module:
First, it's spatially-adaptive, where different kernels are applied for different locations.
Second, instead of embedding hand-craft static pooling like max or average pooling,
our module dynamically generates the kernel based on the semantic similarity of features.

% 作用
For pixel-wise prediction tasks like image translation and semantic segmentation,
the structure of output image needs to align with input image.
Suffering from the low-resolution bottleneck, encoder-decoder architectures lacks fine-grained structure alignment.
Traditional methods like U-net often use skip connections to alleviate this problem.
The skip connection propagates encoder's high-resolution information to upsampled output with concatenation for precise localization.
However, it only strengthens the spatial alignment instead of semantic alignment.
By contrast, our proposed SSA repairs the semantic misalignment by relocating each local patch of upsampled outputs
with the high-resolution feature maps of encoder(The influence of the patch width will be further discussed in ablation study.).

Besides, the SSA has the edge-preserving smoothing characteristic especially suitable for cartoonization task.
For features within the same semantic region, the kernel of semantic candidates tends to be an average kernel,
which helps render smooth surface in this region.
For features near the border of a semantic region, the kernel will assign dominant weight to locations within the border.
Thus the edges between different objects can be well preserved, which is consistent with cartoon image's property, as shown in Figure~\ref{fig:intro}.

\subsection{Perceptual Motion Consistency}

Traditional methods often utilized warping error with optical flow to inhabit temporal inconsistency between output frames~\cite{Chen2017b,Park2019c,Kim2019d}.
The formulation of warping error is:
\begin{equation}
    \mathcal{L}_{\text {warp}}  = \| t^i - W_{{i-1}, i} (t^{i-1}) \|
\end{equation}
where $W_{{i-1}, i}$ denotes the optical flow that warps pixels of $s^{i-1}$ to $s^i$.

However, there are three limitations of the pixel-wise temporal regularization.
First, optical flow describes the spatial variation of pixel but ignores its value variation.
Actually, the brightness and color appearances of the same object might be different on consecutive frames due to illumination effects.
As a consequence, warping error tends to keep the same value corresponding to last frame's output, even when the input value has changed.
Second, images in different domains have different properties in appearances and textures,
optical flow estimated in the photo domain is hard to align with that in cartoon domain, which is illustrated in Figure~\ref{fig:visual}.
Third, optical flow can not handle the disocclusion problem since newly appeared pixels have no correspondence with last frame.

Here, we propose the perceptual motion consistency to formulate a style-independent global-aware motion representation under photo and cartoon domains.
We denote $h$ as the pretrained feature extractor.
Take input video as example, for the extracted feature $h(s^i)_j$ of patch $j$ in current frame $s^i$,
we compute its similarity magnitudes with all features of next frame.
We introduce it as the \textit{spatio-temporal correlative map}, formally:
\begin{equation}
    C_{s^{i}_{j}, s^{i+1}} = \left(h\left({s^i}\right)_j \right)^T \Bigl(h\left({s^{i+1}}\right) \Bigr)
\end{equation}
where $h\left({s^i}\right)_j \in \mathbb{R}^{M \times 1}$, $M$ is the number of channels in feature maps,
$h\left({s^{i+1}}\right) \in \mathbb{R}^{M \times N}$ contains features for all locations of size $N$ in frame $s^{i+1}$,
such that $C_{s^{i}_{j}, s^{i+1}} \in \mathbb{R}^{1 \times N}$ captures the correlation between source patch $j$ with each patch of next frame.

Next, we formulate the perceptual motion representation between consecutive frames as a collection of
spatio-temporal correlative maps:
\begin{equation}
    C_{s^{i}, s^{i+1}} = [C_{s^{i}_0, s^{i+1}}; C_{s^{i}_1, s^{i+1}} ; ...; C_{s^{i}_N, s^{i+1}}] \in \mathbb{R}^{N \times N}
\end{equation}

After that, we calculate the multi-level spatio-temporal correlative maps corresponding to extracted features at different scales between source and target domain.
Here cosine loss is used to restrict the perceptual motion consistency:
\begin{equation}
    \mathcal{L}_{\text {motion}} = \|1 - \cos(C_{s^{i}, s^{i+1}}, C_{t^{i}, t^{i+1}})  \|_1
\end{equation}

The spatio-temporal correlative maps model the motion as correspondence of semantic features in a global context.
% Compared to optical flow
It's more robust under domains with different appearances and styles by exploiting the similarity measurement of high-level features.
Besides, it can handle the problem of disocclusion by leveraging the global relationships of newly appeared objects.

\subsection{Loss Functions}
Besides the perceptual motion loss,
the cartoonization effect of generated images is also optimized with adversarial loss.
Here, we model three representations extracted from images to optimize the generated cartoon images,
including the surface representation that contains smooth surface of cartoon images,
the structure representation that refers to flattened global content,
the texture representation that reflects textures and details in cartoon images.

To derive surface representation, guided filter $\mathcal{F}_{gf}$~\cite{He2013} is adopted for edge-preserving filtering.
It removes the textures and details of input image with the help of a guided image, which can be the input image itself.
A discriminator $D_s$ is leveraged to determine whether the structure representations of output images are similar to that of cartoon images.
Let $I_p$ denote the input photo and $I_c$ indicate the reference cartoon images, the surface loss is
\begin{equation}
    \begin{aligned}
        \mathcal{L}_{\text {surface}}(G & , D_{s}) =\log D_{s}(\mathcal{F}_{gf}(\boldsymbol{I}_{c}, \boldsymbol{I}_{c}))  \\
                                        & +\log (1-D_{s}(\mathcal{F}_{gf}(G(\boldsymbol{I}_{p}), G(\boldsymbol{I}_{p}))))
    \end{aligned}
\end{equation}

\begin{figure*}[t]
    \centering
    \includegraphics[width=\linewidth]{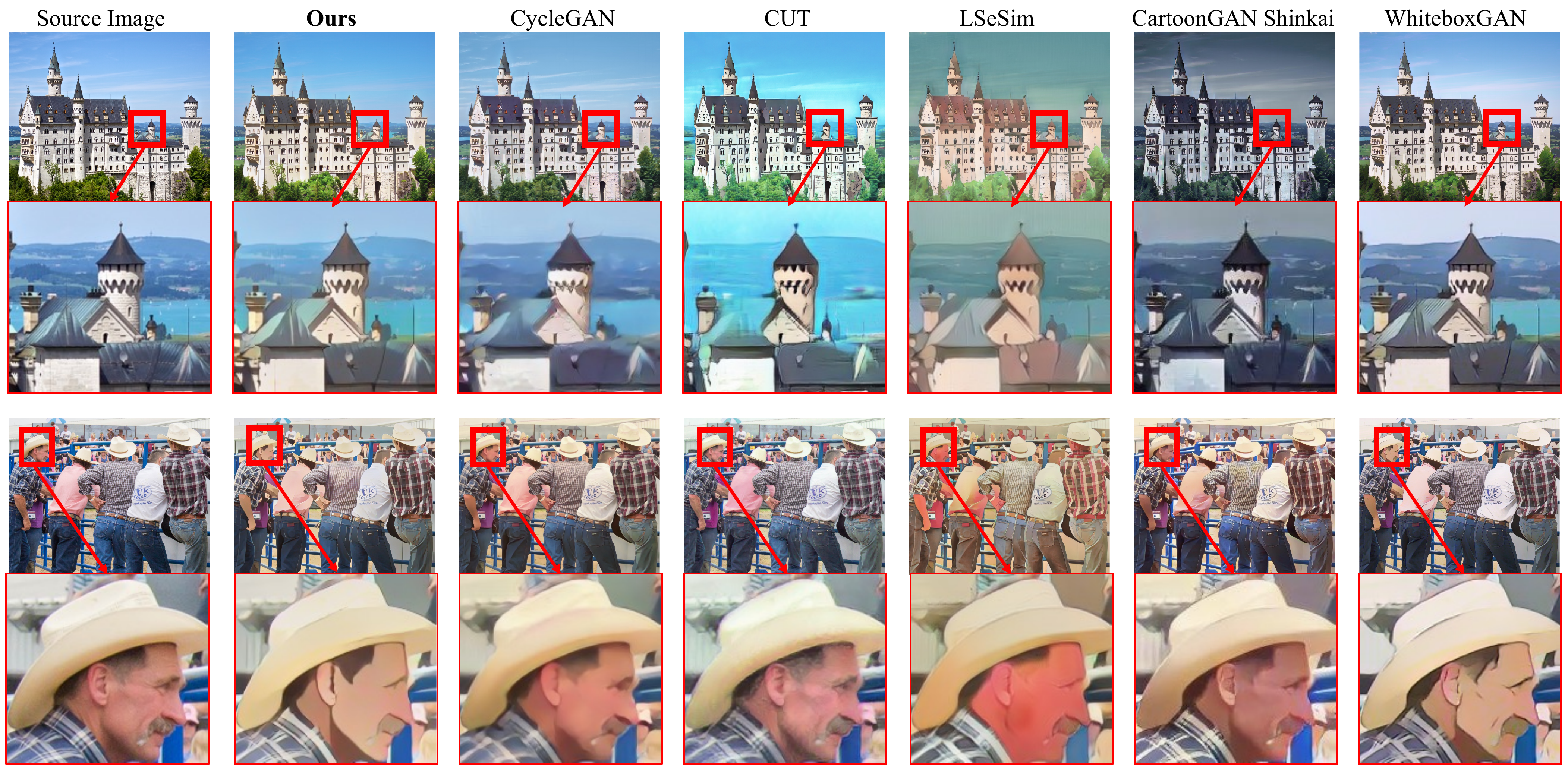}
    \caption{Qualitative comparison on cartoon effects, the first and third rows show the source image and generated cartoon images of different methods.
        The second and fourth rows refer to the magnified details in the red box of images.}
    \label{fig:photo}
\end{figure*}

To derive structure representation, a superpixel algorithm~\cite{felzenszwalb2004efficient} is deployed to segment images into separate semantic consistent regions.
After that, each region is filled with a corresponding color, which generates the structure representation $\mathcal{F}_{st}$,
The structure loss is:
\begin{equation}
    \mathcal{L}_{\text {structure }}=\|V G G(G(\boldsymbol{I}_{p}))-V G G(\mathcal{F}_{s t}(G(\boldsymbol{I}_{p})))\|
\end{equation}

To derive texture representation, we convert images into grayscale.
Another discriminator $D_t$ is used to distinguish the distribution of cartoon images from generated images.
And the texture loss is formulated as:
\begin{equation}
    \begin{aligned}
        \mathcal{L}_{\text {texture }}(G, D_{t}) & =\log D_{t}(\mathcal{F}_{gray}(\boldsymbol{I}_{c}))        \\
                                                 & +\log (1-D_{t}(\mathcal{F}_{gray}(G(\boldsymbol{I}_{p}))))
    \end{aligned}
\end{equation}

Besides, content loss is utilized to regularize the structure consistency between photo and cartoon images,
\begin{equation}
    \mathcal{L}_{\text {content }}=\left\|VGG\left(G\left(\boldsymbol{I}_{p}\right)\right)-VGG\left(\boldsymbol{I}_{p}\right)\right\|
\end{equation}

The total variation loss is incorporated to encourage the smoothness of generated images,
\begin{equation}
    \mathcal{L}_{t v}=\frac{1}{H * W * C}\left\|\nabla_{x}\left(G\left(\boldsymbol{I}_{p}\right)\right)+\nabla_{y}\left(G\left(\boldsymbol{I}_{p}\right)\right)\right\|
\end{equation}

Finally, the objective of full model is formulated as the summation of adversarial loss, perceptual motion loss, content loss and total variation loss:
\begin{equation}
    \begin{aligned}
        \mathcal{L}_{\text {total }} & =\lambda_{1} \mathcal{L}_{\text {surface }}+\lambda_{2}  \mathcal{L}_{\text {texture }}  +\lambda_{3}  \mathcal{L}_{\text {structure }} \\
                                     & +\lambda_{4}  \mathcal{L}_{\text {content }}+\lambda_{5}  \mathcal{L}_{tv} + \lambda_{6} \mathcal{L}_{\text{motion}}
    \end{aligned}
\end{equation}

\section{Experiment}
\subsection{Experimental Setup}

\begin{figure*}[h]
    \centering
    \includegraphics[width=\linewidth]{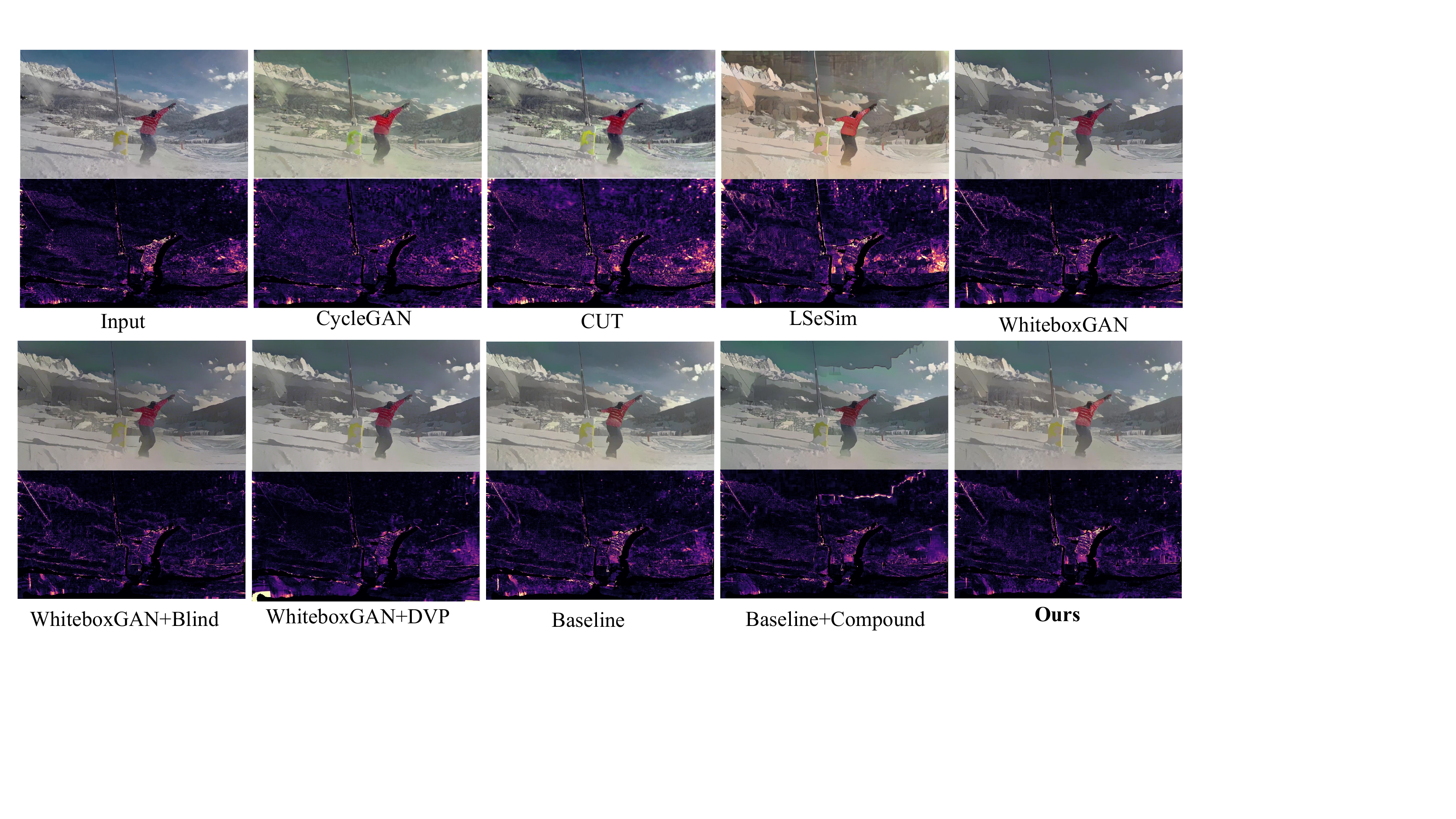}
    \caption{Qualitative comparison on temporal consistency. The first and third rows show source image and generated cartoon images of different methods.
        The second and fourth rows show the heat maps of warping error that indicate the differences between two adjacent video frames.
        Please refer to the supplementary materials for a video demonstration.
    }
    \label{fig:error_maps}
\end{figure*}

\noindent \textbf{Dataset.}
For real-world photos, we adopt 10000 human face images from FFHQ dataset and 6227 landscape images from CycleGAN dataset ~\cite{Zhu2017b}.
For cartoon images, we use images from WhiteboxGAN ~\cite{Wang2020a}, including 10000 images of cartoon faces from P.A.Works, Kyoto animation and
14615 images from cartoon movies produced by Shinkai Makoto, Hosoda Mamoru, and Miyazaki Hayao.
We apply random affine transformations on photo images to imitate input consecutive video frames.
For test set, we use 1541 real-world images from ~\cite{Zhu2017b,Chen2020b} and 1583 cartoon images from above cartoon movies.
During training, all images are resized to $256 \times 256$ resolution.

\noindent \textbf{Implementation Details.}
We implement our model using Pytorch Lightning~\cite{falcon2019pytorch}.
% generator
Our generator consists of two downsampling layers, four residual blocks and two upsampling layers. LeakyReLU is used as activation function.
The local patch width of SSA is set as $R=3$.
% discriminator
Patch discriminator with spectral normalization~\cite{miyato2018spectral} is adopted to identify each patch's distribution.
% 1. Adam, learning rate, betas
We use Adam~\cite{Kingma2015} optimizer with momentums 0.5 and 0.99. The learning rate is set to 0.0002.
% 2. loss weight
The loss weight are set as $\lambda_1 = 0.1, \lambda_2 = 1, \lambda_3 = 200,\lambda_4=200, \lambda_5=20000, \lambda_6=0.1$.
% 3. temporal loss

\begin{table}[t]
    \begin{tabular}{ccc}
        \toprule
        Methods             & FID to Cartoon              & FID to Photo  \\ \hline
        CycleGAN            & 99.39                       & 99.32                     \\
        CUT                 & 93.07                       & 105.64                    \\
        LSeSim              & 95.26                       & 87.66                     \\
        CartoonGAN  Shinkai & 98.46                       & 70.86                     \\
        CartoonGAN Paprika  & 111.40                      & 136.12                    \\
        CartoonGAN Hosoda   & 98.07                       & 136.12                    \\
        CartoonGAN Hayao    & 107.25                      & 119.81                    \\
        WhiteboxGAN         & 90.98                       & 50.69                     \\
        Ours                & \textbf{87.96}              & \textbf{41.11}            \\
        \bottomrule
    \end{tabular}
    \caption{Performance comparison of different methods.
        Smaller FID values indicates closer distance between generated images and reference images.}
    \label{table:fid}
\end{table}

\subsection{Evaluation of Cartoon Effects}
In quantitative experiments, following recent works ~\cite{Wang2020a,park2020contrastive}, we employ the Frechet Inception Distance (FID)~\cite{Heusel2017a}
to evaluate the quality of generated images.
% FID measures divergence of distribution between generated images and reference images.
% We separately calculate the output images' FID compared to photos and cartoon images.
The FID to cartoon images reflects the quality of cartoon style effects,
and the FID to photo indicates content consistency between input photo and generated images.
We compare with five SOTA methods, including unsupervised image translation methods CycleGAN~\cite{Zhu2017b}, CUT~\cite{park2020contrastive}, LSeSim~\cite{Zheng2021}
and image cartoonization methods CartoonGAN~\cite{Chen2018a} and WhiteboxGAN~\cite{Wang2020a}.

The quantitative results are shown in Table \ref{table:fid}.
We can observe that our model has the lowest FID to cartoon images.
Besides, our model surpasses previous methods in FID to photo images by a large margin(9.58).
This benefits from that our model can better preserve the structure information in source image, which proves the effectiveness of our SSA module.

% 图片结果
The qualitative results are shown in Figure \ref{fig:photo}.
We find that the generated images of our method depict vivid cartoon styles.
First, from the aspects of color style, our result displays brighter lightness and higher image contrast without damaging the overall style of source image.
Other SOTA methods like CUT, LSeSim and CartoonGAN generate images with severe color variation.
Second, our method removes negligible details and presents a smoother surface for each semantic region.
By contrast, compared methods especially for CycleGAN, CUT and CartoonGAN still keep original high-frequency information.
Third, with the help of SSA module, our method highly abstracts source image and preserves semantic-aware consistent structure.
Compared with the previous best model WhiteboxGAN, our method achieves more natural structure abstraction as shown by the man face in the fourth row.

\subsection{Evaluation of Temporal Consistency}
To evaluate the coherence of generated videos, we calculate the widely-used warping error as ~\cite{Lei2020}.
For each output frame $t^i$ of a sequence, we calculate its warping error with frame $t^{i-1}$ for short-term consistency:
\begin{equation}
    E_{short}= \frac{1}{N-1} \sum_{i=2}^{N} \| M \circ (W_{s^{i} \rightarrow s^{i-1}}(t^{i})-t^{i-1}) \|_1
\end{equation}
where $W_{s^{i} \rightarrow s^{i-1}}$ is the backward optical flow between $s^{i-1}$ and  $s^i$,
$M$ is the corresponding occlusion mask.
We also calculate the warping error of each frame with frame $t_1$ for long-term consistency:
\begin{equation}
    E_{long}=\frac{1}{N-1} \sum_{i=2}^{N} \| M \circ (W_{s^{i} \rightarrow s^{1}}(t^{i})-t^{1}) \|_1
\end{equation}

\begin{table}[t]
    \centering
    \begin{tabular}{ccc}
        \toprule
        Method                       & $E_{short} \downarrow$ & $E_{long} \downarrow$ \\ \hline
        Source Video                 & 0.0532                 & 0.262                 \\
        CartoonGAN                   & 0.0810                 & 0.305                 \\
        LSeSim                       & 0.0691                 & 0.272                 \\
        CycleGAN                     & 0.0607                 & 0.240                 \\
        CUT                          & 0.0718                 & 0.264                 \\
        WhiteboxGAN                  & 0.0670                 & 0.266                 \\
        WhiteboxGAN+Blind            & 0.0610                 & 0.296                 \\
        WhiteboxGAN+DVP              & \textbf{0.0490}        & 0.2487                \\
        \midrule
        Baseline                     & 0.0746                 & 0.273                 \\
        Baseline + SSA               & 0.0643                 & 0.262                 \\
        Baseline + PMC               & 0.0667                 & 0.258                 \\
        Baseline + SSA+Compound Loss & 0.0517                 & 0.250                 \\ % 待更新
        Baseline + SSA+PMC(Ours)     & 0.0526                 & \textbf{0.237}        \\
        \bottomrule
    \end{tabular}
    \caption{Short-term and long-term warping error of different models on temporal consistency.}
    \label{table:temporal}
\end{table}

Following recent works, we evaluate the temporal consistency on DAVIS~\cite{Perazzi2016} dataset.
% which contains 30 videos with variable length and totally 2294 frames. 
The optical flow is estimated with pretrained RAFT model~\cite{Teed2020}.
In addition to aforementioned methods,
we also adopt post-processing models including Blind ~\cite{Lai2018a} and Deep Video Prior(DVP)~\cite{Lei2020}
upon the SOTA image cartoonization method WhiteboxGAN for comparison.

The quantitative results are shown in Table~\ref{table:temporal}.
Our method performs best for long-term temporal consistency and achieves lower short-term warping error than that of source video.
For image translation models, CartoonGAN, LSeSim, CUT and WhiteboxGAN have higher warping error.
For post-processing methods, Blind can decrease the short-term warping error but increase long-term warping error.
DVP achieves the lowest warping error but it requires several minutes to process a single sequence.
Our method has comparable performance on temporal consistency with post-processing methods,
and this proves the capability of PMC to restrict temporal consistency.

The qualitative results are shown in Figure~\ref{fig:error_maps}, where generated cartoon images with the warping error heat map
are illustrated for different methods.
Our method preserves temporal consistency in most regions except for objects with rapid motion.
Image translation methods in the first row show obvious high-frequency errors in the background.
Post-processing methods decrease the warping error a lot but hurt the cartoon effect (e.g. clear edge).
By contrast, our method both enhance the temporal consistency and renders great cartoon style effects.

\subsection{Ablation Study and Analysis}
We conduct experiments on different variants of our model to evaluate the effectiveness of our proposed SSA and PMC module.
We also adopt the optical flow based compound regularization~\cite{wang2020consistent} for comparison with PMC.
The comparison results are shown in Table \ref{table:temporal} and Figure \ref{fig:error_maps}.

\textbf{Effectiveness of SSA}
By introducing SSA, our model decreases the short-term warping error by 0.0112 and the long-term warping error by 0.011.
As shown in generated images, SSA helps remove the obvious strokes generated by baseline model
and better preserve the semantic-aware structure consistency.
This indicates the structure consistency deriving from SSA can also benefit the temporal consistency.

\textbf{Effectiveness on temporal consistency of PMC}
Simply applying PMC on baseline model benefits both short-term and long-term temporal consistency.
By adding PMC regularization upon SSA, our model further decreases the short-term and long-term warping error by 0.0117 and 0.025.
The compound regularization boosts the temporal consistency but it generates artifacts where a long edge segments the background region.
By contrast, the PMC regularization can better restrict the temporal consistency without hurting the quality of generated images.

\begin{figure}[t]
    \centering
    \includegraphics[width=\linewidth]{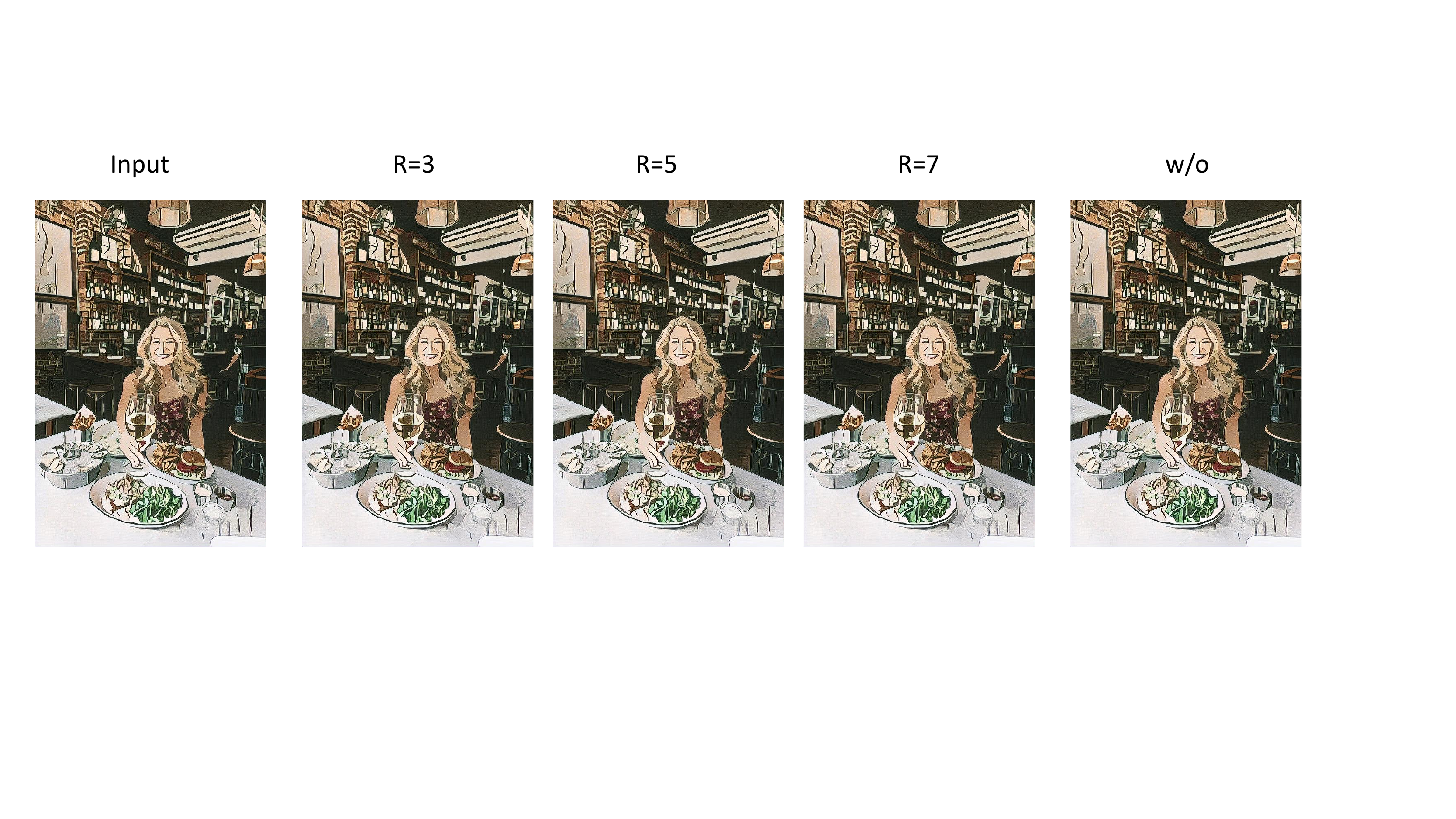}
    \caption{Images generated by our model without SSA and with SSA of different patch widths. Please zoom in to see the details.}
    \label{fig:ablation}
\end{figure}

\textbf{Analysis on the local patch width of SSA}
To explore the relationship between granularity of semantic alignment and local patch width in SSA,
we conduct experiments without SSA and with SSA of different patch widths from 3 to 7.
As shown in Figure~\ref{fig:ablation}, (1) the generated image without SSA demonstrates obvious artifacts and structure inconsistency.
(2) As the increase of patch kernel size, the generator can capture more coarse semantic regions
and the generated images have smoother structure.

\section{Conclusion}
In this paper, we propose a spatially-adaptive semantic alignment framework with perceptual motion consistency to generate
temporal consistent cartoon videos.
The proposed SSA module restores the local shift and deformation of semantic structure, which helps render semantic-aware structure consistent images.
The PMC module builds a style-independent global-aware regularization on perceptual motion consistency to generate coherent cartoon videos.
Experiments and ablation study demonstrate the effectiveness of our proposed modules.

\section*{Acknowledgements}
This work was supported in part by the National Key R\&D Program of China under Grand:2018AAA0102000, National Natural Science Foundation of China: 61771457,61732007,62022083, 
and in part by Youth Innovation Promotion Association of Chinese Academy of Sciences under Grant 2020108, 
CCF-Baidu Open Fund, and CAAI-Huawei MindSpore Open Fund.

\bibliography{aaai22}

\end{document}